# A Visually Attentive Splice Localization Network with Multi-Domain Feature Extractor and Multi-Receptive Field Upsampler

Ankit Yadav, Dinesh Kumar Vishwakarma *Senior Member, IEEE*

*Abstract*—Image splice manipulation presents a severe challenge in today's society. With easy access to image manipulation tools, it is easier than ever to modify images that can mislead individuals, organizations or society. In this work, a novel, "Visually Attentive Splice Localization Network with Multi-Domain Feature Extractor and Multi-Receptive Field Upsampler" has been proposed. It contains a unique "visually attentive multi-domain feature extractor" (VA-MDFE) that extracts attentional features from the RGB, edge and depth domains. Next, a "visually attentive downsampler" (VA-DS) is responsible for fusing and downsampling the multi-domain features. Finally, a novel "visually attentive multi-receptive field upsampler" (VA-MRFU) module employs multiple receptive field-based convolutions to upsample attentional features by focussing on different information scales. Experimental results conducted on the public benchmark dataset CASIA v2.0 prove the potency of the proposed model. It comfortably beats the existing state-of-the-arts by achieving an IoU score of 0.851, pixel F1 score of 0.9195 and pixel AUC score of 0.8989.

## I. INTRODUCTION

IN an age where the alteration of digital content has become effortlessly easy, the authenticity and integrity of visual information have faced substantial hurdles. Image splicing is a very deceitful method of modification that is prevalent in the digital world. Image splicing is the covert practice of merging different visual components from diverse origins to produce an artificial image, typically tricking or misleading onlookers. The ubiquity of image editing tools and the ease of access to digital media have amplified the risks associated with image splicing, exacerbating the propagation of misleading information across various domains, including journalism, forensic analysis, and social media. Consequently, the urgent need for robust and efficient methods to detect these manipulations has become increasingly apparent.

(Corresponding author: Ankit Yadav). Ankit Yadav and Dinesh Kumar Vishwakarma are with the Department of Information Technology, Delhi Technological University, Bawana Road, Delhi-110042, India. (email: ankit4607@gmail.com; dvishwakarma@dtu.ac.in;).

Traditional handcrafted feature-based methods have proven ineffective especially for accurate localization of forgery regions within an image. However, the growth of deep-learning models and the rise of visual attention mechanisms have solved this problem to a great extent.

Several research works are dedicated for this purpose. Zhang et al. [1] propose a novel multi-task squeeze and excitation network (SE-Network) developed explicitly for splicing localization in images. The network incorporates two streams, namely label mask and edge-guided, inside a convolutional encoder-decoder architecture. The method effectively utilises picture edges, label masks, and mask edges to provide thorough supervision. It combines low-level and high-level feature maps and incorporates squeezing and excitation attention modules.

Sun et al. [2] introduce an Edge-enhanced Transformer (ET) for precise tampered region localization in images by effectively integrating splicing edge clues into a two-branch edge-aware transformer. It addresses the challenge of accurately detecting tampered regions without false alarms by enhancing forgery features with edge information and incorporating a feature enhancement module.

Huang et al. [3] develop DS-UNet, a new dual-stream UNet architecture particularly developed for the purpose of detecting picture manipulation and accurately identifying the exact locations of manipulated regions. The DS-UNet algorithm uses an RGB stream to roughly determine the location of tampered items and a Noise stream to accurately detect them. It combines both streams in a hierarchical manner to detect tampered objects of different sizes.

Deng et al. [4] propose MSD-Nets to solve the $QF_1 > QF_2$ case problem by using a *discriminative module* comprising of a CNN trained on $QF_1 > QF_2$ examples exclusively. Multi-scale features are first extracted from DCT histograms which are then fused in a weighted manner. Then, the discriminative module is utilized for the challenging scenario of $QF_1 > QF_2$. Localization results prove the high robustness of the proposed architecture.

The major contributions of this manuscript are:
- Proposed a novel splice localization network with

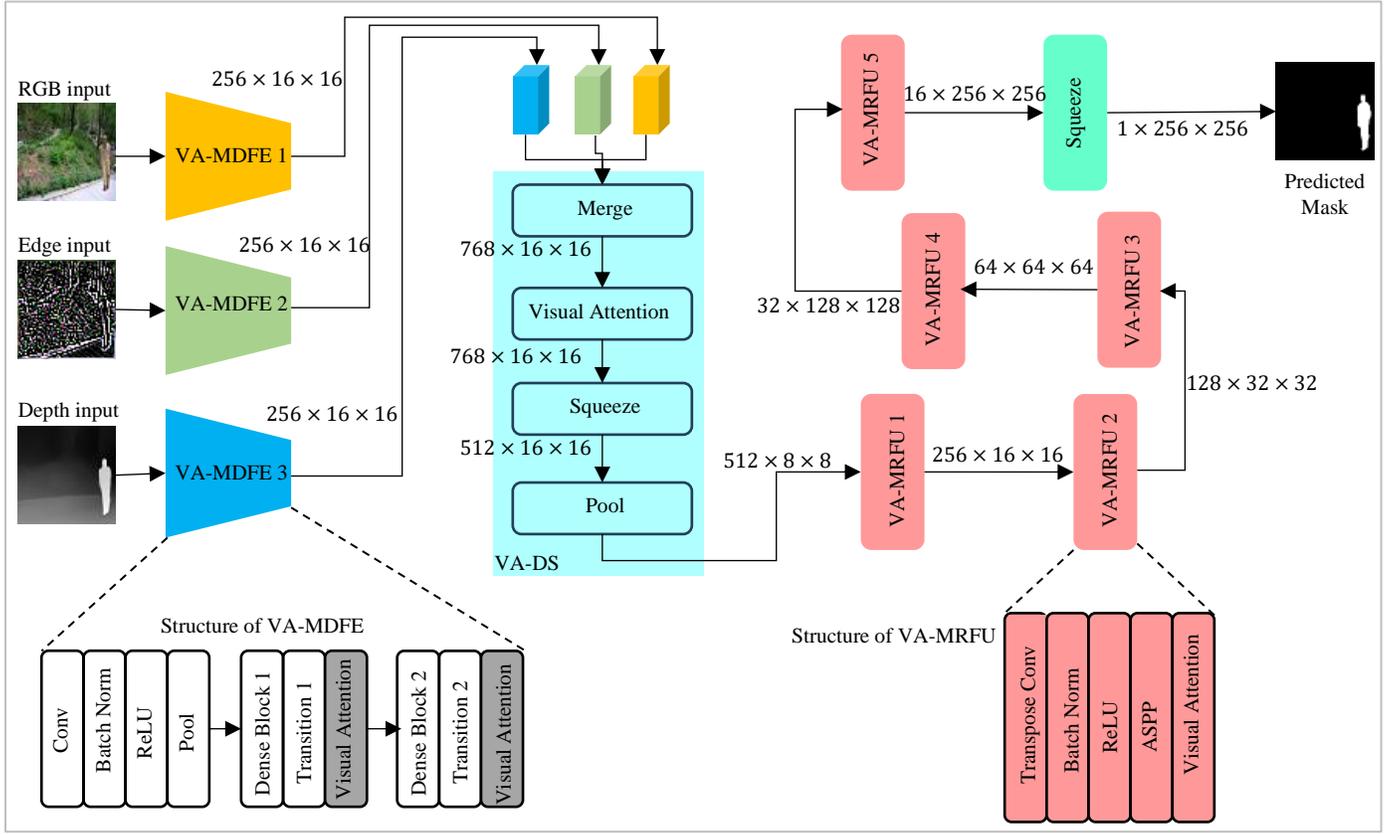

Fig. 1 The structure of the proposed splice localization model. VA-MDFE stands for "visually attentive multi-domain feature extractor". VA-DS stands for "visually attentive downsampler". VA-MRFU stands for "visually attentive multi-receptive field upsampler".

"visually attentive multi-domain feature extractor" (VA-MDFE) and "visually attentive multi-receptive field upsampler" (VA-MRFU).

- The VA-MDFE extracts multi-domain features from the RGB, edge and depth domain aided by the visual attention layer.
- The VA-MRFU is responsible for upsampling features by using multiple receptive field-based convolution operations. This is achieved by varying the dilation rate of convolutional kernels.
- Experimental results on the public benchmark dataset CASIA v2.0 prove that the proposed model outperforms several state-of-the-art methods by achieving an IoU score of 0.851, pixel F1 score of 0.9195 and pixel AUC score of 0.8989.

The rest of the paper is organized as follows. Section II presents the proposed splice localization model. Section III describes the experimental setup. Section IV presents the performance of the proposed model. Section V gives the conclusion, and Section VI includes the references cited in this research work.

## II. PROPOSED ARCHITECTURE

This section presents the architecture of the proposed model.

*A. Visually Attentive Multi-Domain Feature Extractor (VA-MDFE)*

This section describes the visually attentive multi-domain feature extractor of the proposed model. Specifically, a baseline model aided with visual attention extracts features from the input image's RGB, edge and depth domain.

The design proposed in [5] serves as the foundational architecture in this paper. The system promotes the reuse of features by allowing a layer to access feature maps from all preceding levels, hence enhancing the overall flow of information throughout the network., as shown in Eq. 1:

$$\mathfrak{y}_L = \emptyset_L([\mathfrak{y}_0, \mathfrak{y}_1, \mathfrak{y}_2, \mathfrak{y}_3 \ldots \ldots \mathfrak{y}_{L-1}]) \quad (1)$$

The baseline contains four dense blocks and three transition blocks. The last two dense blocks and transition blocks are discarded for computational efficiency.

Each of the two remaining dense blocks of this baseline is appended with a visual attention layer [6] that employs a three-branch design using a 'rotational' module to detect distinct elements in three different orientations. This approach ensures efficient computation with a minimal parameter count of just 300. Two branches handle the input of shape $\mathcal{X} \in \mathbb{R}^{H \times W \times C}$ and employ a distinct Z-pool mechanism to extract important channel characteristics along the height $H$ and $W$ dimensions. Meanwhile, the a traditional spatial attention is computed in the third branch.

Three instances of VA-MDFE is used to create a three-branch architecture for multi-domain feature extraction as shown in Fig. 1. The three input domains are RGB, edge and depth.

## B. Visually Attentive Downsampler (VA-DS)

The proposed model contains a novel "visually attentive downsampler" (VA-DS). VA-DS is responsible for aggregating features from each domain. VA-DS has two stages namely 'merge' and 'downsample'. The following equation describes the operations of VA-DS:

$$f_{down} = \mathcal{P}(\mathcal{Sq}(\mathcal{VA}(\mathcal{M}(f_i)))) \quad (2)$$

Here $f_i$ refers to the input features from each of the three domains. $\mathcal{M}(.)$ stands for the merge operation and it is achieved by concatenating features along the channel dimension. $\mathcal{VA}(.)$ refers to the visual attention layer applied after merging to highlight important regions within the features fused from multiple domains. $\mathcal{Sq}(.)$ means 'squeeze' operation, which reduces the number of channels via $1 \times 1$ convolution. $\mathcal{P}(.)$ is the pool operation to reduce the spatial resolution of features

## C. Visually Attentive Multi-Receptive Field Upsampler (VA-MRFU)

The proposed model contains a novel "visually attentive multi-receptive field upsampler" (VA-MRFU). VA-MRFU is responsible for upsampling the multi-domain features extracted from VA-MDFE. The following equation can describe this module:

$$f_{up,i} = \mathcal{VA}(\mathcal{ASPP}(\mathcal{Re}(\mathcal{BN}(\mathcal{TC}(f_{multi}))))) \quad (3)$$

Here $f_{up,i}$ are the upsampled features from one VA-MRFU module. $f_{multi}$ represents the multi-domain features extracted from VA-MDFE. $\mathcal{TC}(.)$ stands for transpose convolution with kernel size and stride taken as 2. $\mathcal{BN}(.)$ and $\mathcal{Re}(.)$ represent batch normalization and ReLU activation, respectively.

The multi-receptive field mechanism has been implemented via the atrous spatial pyramid pooling module represented here as $\mathcal{ASPP}(.)$. Specifically, it is a three-branch architecture that performs convolution over the input with varying receptive fields. This is achieved by varying the 'dilation' parameter of the convolution operation to change the receptive fields without increasing the computational cost. Dilation rates of 2, 3 and 4 are used in this experiment in each VA-MRFU module.

Finally, $\mathcal{VA}(.)$ represents the attention layer similar to the ones used in the above modules.

## III. EXPERIMENTAL SETUP

This part outlines the specific experimental parameters employed in this study to assess the efficacy of the proposed approach.

### A. Dataset

The CASIA v2.0 is a challenging image tampering dataset containing 7491 original and 5123 tampered images [7]. The manipulated images are created through various techniques like copy-move, splicing, and removal. The manipulated images aim to simulate real-world tampering scenarios, providing researchers with comprehensive data to develop and evaluate algorithms for detecting and analyzing image forgeries.

### B. Preprocessing, Hyperparameters, Hardware & Loss Function

All images are resized to the resolution of $256 \times 256$. Each pixel value is normalized to the range of [0,1].

All experiments are run for 20 epochs. Adam optimizer is used for weight updation. The learning rate is initialized to 0.0001 and is decayed linearly by 10% after every epoch.

Two 24GB NVIDIA RTX A5000 GPUs are run in parallel for this experiment.

The 'focal loss' from [8] has been used to train the proposed model. This loss is ideally suited for the background-foreground class imbalance problem in object detection scenarios.

### C. Evaluation Metrics

The following metrics have been to measure the localization capabilities of the proposed model.

The *Intersection over Union* (IoU) quantifies the degree of overlap between the predicted and ground truth masks by calculating the ratio of the intersection area to the union area of these areas. Higher iou scores suggest superior alignment and accuracy in localization tasks, where a value of 1 signifies complete overlap.

*Pixel-level accuracy* measures the correctness of localised image modifications by determining the proportion of properly identified pixels out of the total number of pixels in a picture. Greater pixel-level accuracy ratings suggest more alignment and precision in localised alterations.

*Pixel-level F1* score is a quantitative measure employed to assess the precision and memory of localised picture alterations at the individual pixel level. The F1 score quantifies the trade-off between correctly identified manipulated pixels (true positives) and the accuracy of the localised manipulation compared to the ground truth annotations. Higher F1 scores indicate better overall accuracy in capturing manipulated regions.

*Pixel-level AUC*, also known as Area Under the Curve, is a quantitative metric utilised to evaluate the effectiveness of detecting localised image modification at different thresholds. It assesses the balance between the rate of correctly identified manipulated areas and the rate of incorrectly identified manipulated regions at the individual pixel level. A greater pixel-level AUC signifies superior differentiation between manipulated and non-manipulated areas, demonstrating the efficacy of the detection system.

## IV. EXPERIMENTAL RESULTS

This section presents the experimental results obtained for the proposed model on the benchmark datasets.

### A. Performance on Benchmark Dataset CASIA v2.0

Fig. 2 presents the performance on the CASIA v2.0 dataset in terms of the IoU, pixel accuracy, pixel F1 and pixel AUC scores. The proposed model achieves excellent scores, highlighting its strong localization capability.

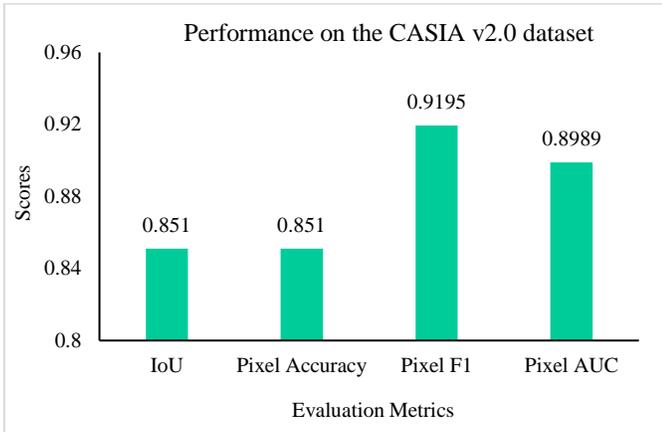

Fig. 2 Performance of the proposed model on CASIA v2.0 dataset. All metrics are pixel-level except IoU.

*B. Comparison Against the State-of-the-Arts*

Table I compares the proposed model against the existing state-of-art methods. The proposed model easily outperforms other methods, demonstrating its superiority.

Table I Comparison of the Proposed Model against existing state-of-the-art methods. The best scores are highlighted in bold.

| Methods | IoU | Pixel F1 | Pixel AUC |
|---|---|---|---|
| Zhang et al. [1] | 0.7139 | 0.7653 | - |
| Sun et al. [2] | 0.5157 | 0.6805 | - |
| Huang et al. [3] | - | 0.6100 | 0.749 |
| Nazir et al. [9] | - | 0.8469 | - |
| Yin et al. [10] | - | 0.5840 | 0.8950 |
| RRU-Net [11] | 0.4752 | 0.5333 | - |
| ManTra-Net [12] | 0.1261 | 0.2009 | - |
| Chen et al. [13] | 0.4386 | 0.6097 | - |
| **Proposed Model** | **0.8510** | **0.9195** | **0.8989** |

*C. Qualitative Analysis*

This section presents a visual comparison of the forgery masks predicted by the proposed model.

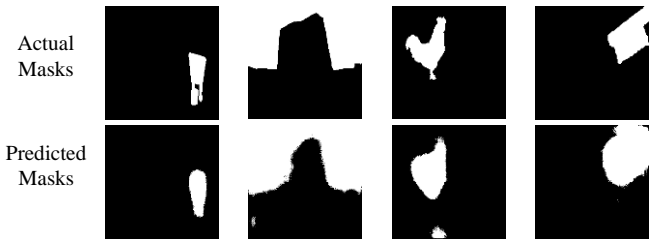

Fig. 3 A comparison of the actual and predicted masks from the proposed model.

*D. Ablation Study*

This section presents an ablation study of the proposed model. Specifically, each of the individual domains are evaluated separately. This means that the proposed model is compared against single domain feature extractors with RGB, edge and depth images as input.

Table II Comparison of single-domain vs the proposed multi-domain model.

| Methods | IoU | Pixel Accuracy | Pixel F1 | Pixel AUC |
|---|---|---|---|---|
| RGB Only | 0.5494 | 0.7245 | 0.6201 | 0.637 |
| Edge Only | 0.5164 | 0.614 | 0.652 | 0.5497 |
| Depth Only | 0.4633 | 0.6681 | 0.6332 | 0.5804 |
| **Proposed Model** | **0.851** | **0.851** | **0.9195** | **0.8989** |

Table II presents the performance of single-domain approaches against the proposed model. Specifically, each domain from the RGB, edge and depth is used as an individual feature extractor in the single domain models.

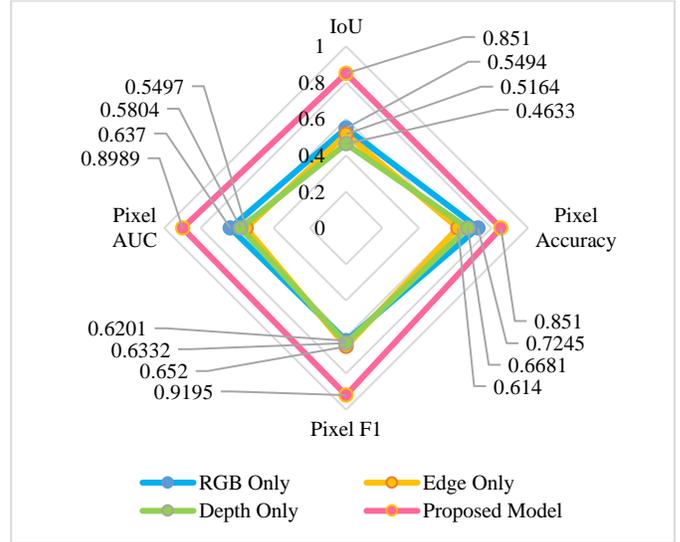

Fig. 4 A visual comparison of single domain (RGB, edge and depth) against the proposed multi-domain feature extractor. The proposed model (pink) achieves the highest IOU, pixel accuracy, pixel F1 and pixel AUC score.

Fig. 4 visually compares the ablation study conducted where single domain architectures are compared against the proposed model. The proposed model (pink) outperforms the individual domains of RGB (blue), edge (orange) and depth (green) across all metrics, namely IoU, pixel accuracy, pixel F1 and pixel AUC. This clearly states the benefit of having a multi-domain feature extractor.

## V. CONCLUSION

In this manuscript, a novel image splice localization network is proposed. The proposed model contains a novel "visually attentive multi-domain feature extractor" that extracts attentional features from the RGB, edge and depth domain. A novel "visually attentive multi-receptive field upsampler" is responsible for the upsampling of features using multiple receptive field-based convolution operation. Experimental results on the CASIA v2.0 public benchmark dataset prove the potency of the proposed model as it easily beats the existing research approaches of splice localization.